\documentclass[letterpaper, 10 pt, conference]{ieeeconf} 
\usepackage{times}
\usepackage{color, colortbl, xcolor}
\newcommand{\hdyn}{f_H}
\newcommand{\hstate}{x_H}
\newcommand{\hderiv}{\dot{x}_H}
\newcommand{\hctrl}{u_H}
\newcommand{\cset}{\mathcal{U}}
\newcommand{\hxdim}{{n_H}}
\newcommand{\hudim}{{m_H}}

\newcommand{\param}{\lambda}
\newcommand{\belief}{P}
\newcommand{\pt}{\belief^{\tvar}}

\newcommand{\postpt}{\belief^{\tvar}_{+}}

\newcommand{\ptparam}{\pt(\param)}

\newcommand{\jointstate}{z}
\newcommand{\jointdyn}{f}
\newcommand{\betaintrdyn}{k}

\newcommand{\rdyn}{f_R}
\newcommand{\rstate}{x_R}
\newcommand{\rderiv}{\dot{x}_R}
\newcommand{\rctrl}{u_R}
\newcommand{\rxdim}{n_R}
\newcommand{\rudim}{{m_R}}
\newcommand{\rgoal}{{g_R}}

\newcommand{\tvar}{t}
\newcommand{\R}{\mathbb{R}}
\newcommand{\state}{\jointstate}
\newcommand{\targetfunc}{l}
\newcommand{\targetset}{\mathcal{L}}
\newcommand{\vfunc}{V}
\newcommand{\frs}{\mathcal{V}} % forward reachable set
\newcommand{\frspos}{\mathcal{K}} % FRS in the position space
\newcommand{\approxfrspos}{\tilde{\mathcal{K}}} % Approx. FRS in the position space
\newcommand{\proj}{\Pi}
\newcommand{\ham}{H}
\newcommand{\tdummy}{\tau}
\newcommand{\tmin}{T_{\text{min}}}
\newcommand{\tmax}{T_{\text{max}}}

\newcommand{\horizon}{T}

\newcommand{\finaltime}{\bar{T}}

\newcommand{\frsname}{\text{BA-FRS}}

\usepackage{multicol}
\usepackage[bookmarks=true]{hyperref}

\usepackage{amsfonts}
\usepackage{algorithmic}
\usepackage[ruled,vlined,titlenotnumbered]{algorithm2e} 
\usepackage{amsmath}
\usepackage{dsfont}
\usepackage{subfigure}
\usepackage{multicol}
\usepackage[pdftex]{graphicx}   
\usepackage{graphicx}

\graphicspath{{./figs/}}    
\DeclareGraphicsExtensions{.pdf,.png}  

\usepackage[textfont=md,font=footnotesize]{caption}

\usepackage[left=55pt, right=55pt, top=55pt, bottom=58pt]{geometry} 

\IEEEoverridecommandlockouts
\usepackage{balance}

\usepackage[%
    style=numeric-comp,
    sorting=none,
    backend=biber,
    sortcites=true,
    doi=false,
    firstinits=true,
    hyperref,
    isbn=false,
    eprint=false,
    maxcitenames=3, %How many people before *choosing* to contract with [et al.]
    minbibnames=3, %How many authors *at least*  before *inserting* the [et al.]
    maxbibnames=4, %How many authors *at most*  before *inserting* the [et al.]
    block=none]
    {biblatex}
    
    \renewbibmacro{in:}{}
    \AtEveryBibitem{%
  	\clearlist{language}%
  	\clearfield{pages}%
	}
\bibliography{references}
\usepackage{lipsum}

\newcommand\blfootnote[1]{%
  \begingroup
  \renewcommand\thefootnote{}\footnote{#1}%
  \addtocounter{footnote}{-1}%
  \endgroup
}
\definecolor{planning_color}{RGB}{69, 174, 254}    
\definecolor{prediction_color}{RGB}{255, 116, 190}
    
\newcommand{\example}[1]%
{
\textbf{Running example:}
\textit{#1}
}

\usepackage{ifthen}
\newboolean{include-notes}
\setboolean{include-notes}{true}

\newcommand{\adnote}[1]{\ifthenelse{\boolean{include-notes}}{\textcolor{purple}{{A (comment):#1}}}{}}   
\newcommand{\adedit}[1]{\ifthenelse{\boolean{include-notes}}{\textcolor{magenta}{{A (edit):#1}}}{}}   

\begin{document}

% paper title
\title{A Hamilton-Jacobi Reachability-Based Framework \\for Predicting and Analyzing Human Motion for Safe Planning}

\author{Somil Bansal*, Andrea Bajcsy*, Ellis Ratner*, Anca D. Dragan, Claire J. Tomlin}
% \authorblockA{
% Not sure if it's worth explicitly adding the following under the emails:
%\authorrefmark{1}The first two authors contributed equally to this work.
% }

% avoiding spaces at the end of the author lines is not a problem with
% conference papers because we don't use \thanks or \IEEEmembership

% for over three affiliations, or if they all won't fit within the width
% of the page, use this alternative format:
% 
%\author{\authorblockN{Michael Shell\authorrefmark{1},
%Homer Simpson\authorrefmark{2},
%James Kirk\authorrefmark{3}, 
%Montgomery Scott\authorrefmark{3} and
%Eldon Tyrell\authorrefmark{4}}
%\authorblockA{\authorrefmark{1}School of Electrical and Computer Engineering\\
%Georgia Institute of Technology,
%Atlanta, Georgia 30332--0250\\ Email: mshell@ece.gatech.edu}
%\authorblockA{\authorrefmark{2}Twentieth Century Fox, Springfield, USA\\
%Email: homer@thesimpsons.com}
%\authorblockA{\authorrefmark{3}Starfleet Academy, San Francisco, California 96678-2391\\
%Telephone: (800) 555--1212, Fax: (888) 555--1212}
%\authorblockA{\authorrefmark{4}Tyrell Inc., 123 Replicant Street, Los Angeles, California 90210--4321}}

\maketitle

%\abnote{Camera-TODO: (1) redo introduction, (2) continualization of beta dynamics, (3) dynamic beta example}

% \abnote{We thought about the computational complexity and came up with the following:
% \begin{itemize}
%     \item $\mathcal{O}_{HJ}(n_x n_y \prod^{n_\lambda-1}_{i=1}n_{P(\lambda_i)}n_t)$
%     \item $\mathcal{O}_{Bayes}(n_x n_y n_\lambda n^{n_t}_\lambda)$
% \end{itemize}
% However, Ellis brought up a good point that the $O_{Bayes}$ could be reduced by a smarter implementation where the probabilities of $\lambda$ are put into the state space as well.
% }

\blfootnote{$^*$Indicates equal contribution. Authors are with EECS at UC Berkeley. Research supported by DARPA Assured Autonomy, NSF CPS VeHICal, SRC CONIX, NSF CAREER award, and a NASA NSTRF.}

\vspace{-1.5em}
\begin{abstract}
Real-world autonomous systems often employ probabilistic predictive models of human behavior during planning to reason about their future motion.
Since accurately modeling human behavior \textit{a priori} is challenging, such models are often parameterized, enabling the robot to adapt predictions based on observations by maintaining a distribution over the model parameters.
Although this enables data and priors to improve the human model, observation models are difficult to specify and priors may be incorrect, leading to erroneous state predictions that can degrade the safety of the robot motion plan. In this work, we seek to design a predictor which is more robust to misspecified models and priors, but can still leverage human behavioral data online to reduce conservatism in a safe way. To do this, we cast human motion prediction as a Hamilton-Jacobi reachability problem in the joint state space of the human and the belief over the model parameters. We construct a new continuous-time dynamical system, where the inputs are the observations of human behavior, and the dynamics include how the belief over the model parameters change. The results of this reachability computation enable us to both analyze the effect of incorrect priors on future predictions in continuous state and time, as well as to make predictions of the human state in the future. We compare our approach to the worst-case forward reachable set and a stochastic predictor which uses Bayesian inference and produces full future state distributions. Our comparisons in simulation and in hardware demonstrate how our framework can enable robust planning while not being overly conservative, even when the human model is inaccurate.  
Videos of our experiments can be found at the project website\footnote{Project website: \href{https://abajcsy.github.io/hallucinate/}{https://abajcsy.github.io/hallucinate/}}.
\end{abstract}

\IEEEpeerreviewmaketitle

\section{Introduction}
\begin{figure} [t!]
    \centering
    \includegraphics[width=0.8\columnwidth]{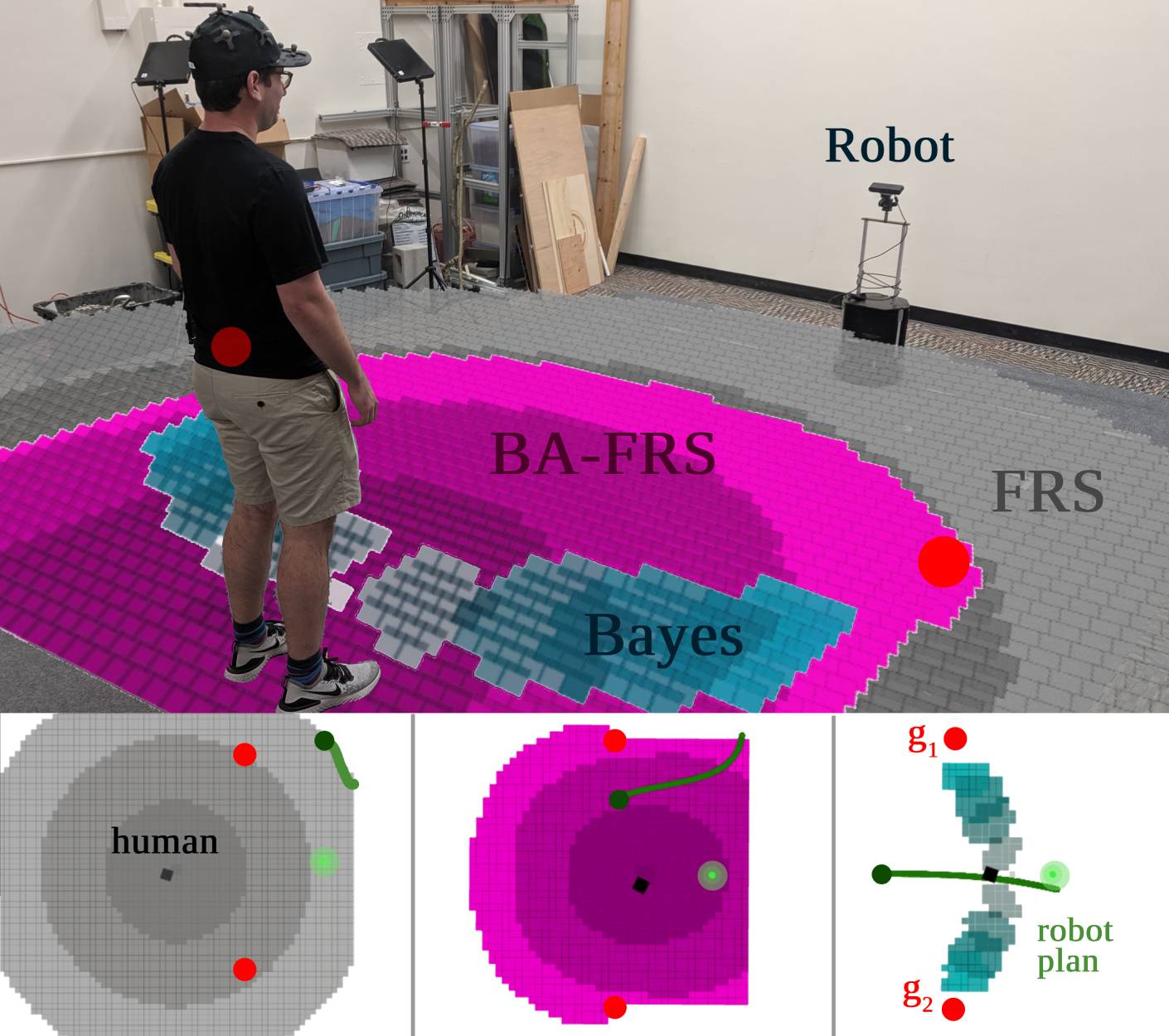}
    \caption{The robot models the human as walking towards one of the two goals shown in red. However, the human actually walks straight in between them. The predictions from our Hamilton-Jacobi reachability framework (center, in magenta) approximates the full probability distribution (right, in teal) while being more robust to model misspecification and less conservative compared to the naive full forward reachable set (left, in grey).}
    \label{fig:front_fig}
    \vspace{-1em}
\end{figure}

Planning around humans is critical for many real-world robotic applications. To effectively reason about human motion, practitioners often couple dynamics models with human decision-making models to generate predictions that are used by the robot at planning time. Such predictors often model the human as an agent maximizing an objective \cite{bai2015intention, baker2007goal, ng2000algorithms, ziebart2009planning, kitani2012activity} or they learn human behavioral structure  from data \cite{Schmerling2017}. When coupled with robot motion planners, these human decision-making models have been successfully used in a variety of domains including manipulation \cite{amor2014interaction, ding2011human, koppula2013anticipating, lasota2015analyzing, hawkins2013probabilistic}, autonomous driving \cite{hawkins2013probabilistic, driggs2018robust}, and navigation \cite{Ma_2017_CVPR, alahi2016social} (see \cite{rudenko2019human} for a survey).

%Such predictive models often encode the relevant structure in human decision-making through model parameters. \ernote{I typically think of ``structure'' as being the mathematical function (e.g. the Boltzmann distribution), which in turn has parameters. Maybe we could re-phrase: .} 
Such predictive models are often parameterized to encode variations in decision-making between different people. However, since modeling human behavior and how people make decisions \textit{a priori} is challenging, predictors can maintain a belief distribution over these model parameters \cite{fisac2018general, lasota2017multiple}.
% This is well-suited for stochastic predictors which generate full-state distributions \cite{ziebart2009planning, bandyopadhyay2013intention, kochenderfer2010airspace} and naturally incorporate the model parameter uncertainty into future state uncertainty. 
This stochastic nature of the predictor naturally incorporates the model parameter uncertainty into future human state uncertainty~\cite{ziebart2009planning, bandyopadhyay2013intention, kochenderfer2010airspace}.
Importantly, once deployed, the robot can observe human behavior and update the distribution over the model parameters to better align its predictive model with the observations. 
    
% However, there are two key challenges with such stochastic predictors. 
% First, there exist limited computational tools to compute these predictions in continuous time and state, which may be important for safety-critical applications. 
% Secondly, and perhaps more crucially, to update the distribution and to generate predictions, stochastic predictors rely on priors and on  observation models. 
% Although these two components enable data and prior knowledge to improve the model, observation models are difficult to specify and the priors may be incorrect. When either is erroneous, then the state predictions will be erroneous as well, and the robot may confidently plan unsafe motions. %Furthermore, it takes non-zero time and observations to correct an incorrect prior -- with traditional stochastic predictors, the robot still relies on these predictions during this time despite the prediction inaccuracies. 

However, there are two key challenges with such stochastic predictors. 
First, to update the distribution and to generate predictions, stochastic predictors rely on priors and on observation models. 
Although these two components enable data and prior knowledge to improve the model, observation models are difficult to specify and the priors may be incorrect. When either is erroneous, then the state predictions will be erroneous as well, and the robot may confidently plan unsafe motions. 
Secondly, there exist limited computational tools to generate these stochastic predictions in continuous time and state, which may be important for safety-critical applications. 

To address the former challenge, researchers in the robust control community have looked into \textit{full forward reachable set} predictors, which compute the set of \textit{all possible} states that the human could reach from their current state \cite{mitchell2005time, driggs2018robust}. 
Unfortunately, because these predictors consider worst-case human behavior, the resulting predictions are overly conservative, do not typically leverage any data to update the human model, and they significantly impact the overall efficiency of the robot when used in practice. 

%To combat some of these issues, the robust optimal control perspective on prediction removes any reliance on data and explicit structure in the human decision-making model. That is, the human is modeled as taking \textit{any} action from a conservative set of controls. Combined with a dynamics model, the predictor generates a \textit{full forward reachable set}, or the set of \textit{all} states that the human could reach from their current state \cite{mitchell2005time, driggs2018robust}. These predictors do not rely on priors or observation models and the control theory community has developed strong theoretic and numerical tools for analyzing the continuous-time evolution of such models. Unfortunately, because these predictors consider worst-case human behavior, the resulting predictions are overly conservative, cannot leverage any data to update the human model, and they significantly impact the overall efficiency of the robot when used in practice. 

%In this work we cast the human motion prediction problem as a reachability problem, which allows us to make a novel connection between the robust optimal control community and the human prediction community.  

In this work we seek a marriage between stochastic prediction and robust control: a predictor which is more robust to misspecified models and priors, but can still leverage human behavioral data online to reduce conservatism in a safe way. To do this, we cast human motion prediction as a Hamilton-Jacobi (HJ) reachability problem \cite{mitchell2005time, abate2008probabilistic, lygeros2004reachability} in the joint state space of the human and the belief over the predictive model parameters.
Observations of the human are treated as continuous ``input'' into our new dynamical system and how the belief over the model parameters changes based on this data are part of the ``dynamics'' of the system. 
However, unlike stochastic predictors, we do not rely on the exact probability of an observation during the prediction. 
Rather, we divide the possible human observations under the current model in two disjoint sets of \textit{likely} and \textit{unlikely} observations, and compute the possible future human states by treating all likely enough observations as equally probable.
%Since we no longer rely on the exact observation probabilities, the resulting predictor ensures safe planning when priors are not correct but can still leverage human behavioral data online to reduce conservatism in a safe way, thus bridging stochastic predictors with the full forward reachable set. 
Since we no longer rely on the exact observation probabilities, the resulting predictor exhibits robustness to incorrect priors, while still leveraging human behavioral data to reduce conservatism online in a safe way. In doing so, our algorithm provides a bridge between stochastic predictors and predicting the full forward reachable set. %(Note that we are not ensuring anything about the robustness of the \emph{planning}, just the robustness of the prediction.)
Additionally, our reachability formulation allows us to leverage the computational tools developed for reachability analysis to predict future human states in continuous time and state \cite{mitchell2004toolbox, chen2013flow}.
% Using this novel reachability formulation, we demonstrate how to bridge stochastic predictors with the full forward reachable set. This allows us to explicitly reason about how incorrect priors can affect our predictions in the future. Additionally, our formulation allows us to leverage the computational tools developed for reachability analysis to predict future human states in continuous time and state \cite{mitchell2004toolbox, chen2013flow}. Our resulting predictor ensures safe planning when priors are not correct but can still leverage human behavioral data online to reduce conservativism in a safe way.

Interestingly, because HJ reachability is rooted in optimal control, our novel formulation also allows us to analyze human models which change online based on data. For example, we can now ask (and answer) questions like ``How long will it take the predictor to reach a desired level of confidence in its human model?'' As long as the robot does not have enough confidence in the human model, this information can be used to plan safe maneuvers. 
Once the confidence in the human model is high enough, the robot can plan more aggressive maneuvers to improve performance.

To summarize, our key contributions in this work are:
\begin{itemize}
    \item a Hamilton-Jacobi reachability-based framework for human motion prediction which provides robustness against misspecified observation models and priors;
    %to design a predictor which is more robust to incorrect priors  when priors are not correct but can still leverage human behavioral data online whenever safe;
    \item a demonstration of how our framework can be used to analyze human decision-making models that are updated online;
    %\ernote{Maybe let's swap the order of the first two-- introducing it as a HJ reachability framework for doing analysis, and then present the predictor that comes out as an additional benefit}
    \item a demonstration of our approach in simulation and on hardware for safe navigation around humans.
\end{itemize}

\section{Problem Statement}
% \abnote{NOTE: should agree on if we are using belief notation $b_t$ or pdf notation $P_t$}
% \begin{itemize}
%     \item Assume known robot goal $\rgoal$, map of static parts of environment environment $\map$
%     \item Want to move $\rstate$ to $\rgoal$ without colliding with human or the static obstacles
%     \item Since we already know the environment map, but we don't know the human occupancies in the future
%     \item We will divide this problem into two steps: prediction and planning
% \end{itemize}
% 
We study the problem of safe motion planning for a mobile robot in the presence of a single human agent.
In particular, our goal is to compute a control sequence for the robot which moves it from a given start state to a goal state without colliding with the human or the static obstacles in the environment. 
We assume that both the robot and human states can be accurately sensed at all times.
Finally, we also assume that a map of the static parts of the environment is known; however, the future states of the humans are not known and need to be predicted online. 
Consequently, we divide the safe planning problem into two subproblems: human motion prediction and robot trajectory planning. 
% \adnote{it's not path if it's timed, and it needs to be timed if you are avoiding a time-moving obstacle} \adnote{I know you're worried about making it roboticsy enough for icra, but i'd say it's obvious enough that you need motion prediction to plan what the robot does (and you can point to many papers for this) that you don't really need to dedicate space in the paper to the robot planning part; it's enough to show an example at the end, where you can write a few sentences about how you did planning with this; i don't feel strongly about this, just a suggestion if space is tight}

\subsection{Human Motion Prediction with Online Updates} \label{sec:human_prediction_setup}
To predict future human states, %we employ a dynamics model of the human's physical states and a model of the human's control policy.
we model the human as a dynamical system with state $\hstate \in \R^\hxdim$, control $\hctrl \in \R^\hudim$, and dynamics
\begin{equation} \label{eqn:human_dynamics}
    \hderiv = \hdyn(\hstate, \hctrl).
\end{equation}
Here, $\hstate$ could represent the position and velocity of the human, and $\hdyn$ describes the change in their evolution over time. 
To find the likely future states of the human, we couple this dynamics model with a model of how the human chooses their actions. In general, this is a particularly difficult modeling challenge and many models exist in the literature (see \cite{rudenko2019human}). 
%For example, the control policy could be hand-engineered based on prior knowledge of how people behave \cite{XYZ}, or learned from data \cite{XYZ}. 
In this work, we primarily consider stochastic control policies that are parameterized by $\param^{\tvar}$:
% wherein at each timestep $\tvar$ the human's control action is sampled from the given control policy which is parameterized by $\param^{\tvar}$:
%
\begin{equation}
    \hctrl^\tvar \sim P(\hctrl^\tvar \mid \hstate^\tvar; \param^{\tvar}).
    \label{eq:control_policy}
\end{equation}
In this model, $\param^\tvar$ can represent many different aspects of human decision-making\footnote{This formulation is easily amenable to deterministic policies where $P(\hctrl^\tvar \mid \hstate^\tvar; \param^{\tvar})$ is the Dirac delta function.}, from how passive or aggressive a person is \cite{sadigh2016information} to the kind of visual cues they pay attention to in a scene \cite{kitani2012activity}. The specific choice of parameterization is often highly problem specific and can be hand-designed or learned from prior data \cite{ziebart2009planning, finn2016guided}. Nevertheless, the true value of $\param^{\tvar}$ is most often not known beforehand and instead needs to be estimated after receiving measurements of the true human behavior. Thus, at any time $\tvar$, we maintain a distribution $\ptparam$ over $\param^{\tvar}$, which allows us to reason about the uncertainty in human behavior online based on the measurements of $\hctrl$. 

\example{We now introduce a running example for illustration purposes throughout the paper.
In this example, we consider a ground vehicle that needs to go to a goal position in a room where a person is walking.
We consider a planar model with state $\hstate = [h_x, h_y]$, control $\hctrl = \theta$, and dynamics $\dot{x}_H = [v_H cos(\theta), v_H sin(\theta)]$.
%
% \begin{equation}
%     \label{eqn:example_human_dynamics}
%     \begin{bmatrix}
%     \dot{h}_x \\
%     \dot{h}_y
%     \end{bmatrix} = \begin{bmatrix}
%                 v_H cos(\theta) \\
%                 v_H sin(\theta) 
%               \end{bmatrix},
% \end{equation}
% \begin{equation}
%     \label{eqn:example_human_dynamics}
%     \dot{x}_H = [v_H cos(\theta), v_H sin(\theta)].
% \end{equation}
%
The model parameter $\param^{\tvar}$ can take two values and indicates which goal location the human is trying to navigate to. 
The human policy for any state and goal is given by a Gaussian with mean pointing in the goal direction and a variance representing uncertainty in the human action:  
    \begin{equation}
       \hctrl^{\tvar} \mid \hstate^{\tvar} \sim 
    \begin{cases}
        \mathcal{N}(\mu_1, \sigma_1^2),  & \text{if } \param^{\tvar} = g_1 \\
        \mathcal{N}(\mu_2, \sigma_2^2),  & \text{if } \param^{\tvar} = g_2
    \end{cases},
    \end{equation}
where $\mu_i = \tan^{-1}\big(\frac{g_i(y) - h_y^{\tvar}}{g_i(x) - h_x^{\tvar}}\big)$ and $\sigma_i = \pi/4$ for $i \in \{1, 2\}$. 
% Here, $\param^{\tvar}$ can take two values and indicates which goal location the human is trying to navigate to.
$(g_i(x), g_i(y))$ represents the position of goal $g_i$.
} 

Since we are uncertain about the true value of $\param^{\tvar}$, we update $\ptparam$ online based on the measurements of $\hctrl^{\tvar}$.
This observed control may be used as evidence to update the robot's prior belief $\ptparam$ about $\param^{\tvar}$ over time via a Bayesian update to obtain the posterior belief:
\begin{equation}
\label{eqn:bayesian_update}
\postpt(\param^{\tvar} \mid \hctrl^\tvar, \hstate^\tvar) = \frac{P(\hctrl^\tvar \mid \hstate^\tvar; \param^{\tvar})\ptparam}{\sum_{\bar{\param}}P(\hctrl^\tvar \mid \hstate^\tvar; \bar{\param})\pt(\bar{\param})}
\end{equation}
Given the human state $\hstate^{\tvar}$, the dynamics $\hdyn$ in \eqref{eqn:human_dynamics}, the control policy in \eqref{eq:control_policy}, and the distribution $\ptparam$, our goal is to find the likely human states at some future time, $\tvar+\tdummy$:
\begin{equation}
    \frspos^{\tvar}_{\epsilon}(\tdummy) = \{\hstate^{\tvar+\tdummy} \mid P(\hstate^{\tvar+\tdummy} \mid \hstate^{\tvar}) > \epsilon\},
\end{equation}
where $\epsilon \ge 0$ is the desired safety threshold and is a design parameter.
When $\epsilon = 0$, we drop the subscript $\epsilon$ in $\frspos^\tvar$.
Using this set of likely human states, our robot will generate a trajectory that at each future time step $\tvar+\tdummy$ avoids $\frspos_{\epsilon}^{\tvar}(\tdummy)$. 

Note that the requirement to compute $\frspos^{\tvar}_{\epsilon}(\tdummy)$ is subtly different from computing the full state distribution, $P(\hstate^{\tvar+\tdummy} \mid \hstate^{\tvar})$.
For computing the full distribution, one can explicitly integrate over all possible future values of $\param$, state, and action trajectories. 
Alternatively, one can use the belief space to keep track of $\ptparam$ over time, and compute the human state distribution using the belief. 
The latter computation can be thought of as branching on future observations, and keeping track of what the belief might be at each future time depending on that observation history and the intrinsic changes in the human behavior.
Our insight is that this latter computation can be formulated as a stochastic Hamilton-Jacobi reachability problem in the joint state space of the human and belief, but that it can be simplified to a deterministic reachability problem.
This not only leads to a more robust prediction of likely future human states when the prior $\ptparam$ is not correct, but also allows us to compute an approximation of $\frspos_{\epsilon}^{\tvar}(\tau)$ with lower computational complexity when the prior is correct.
\subsection{Robot Motion Planning} \label{sec:motion_planning_setup}
We model the robot as a dynamical system with state $\rstate \in \R^\rxdim$, control $\rctrl \in \R^\rudim$, and dynamics $\rderiv = \rdyn(\rstate, \rctrl)$.
%
% \begin{equation} \label{eqn:robot_dynamics}
%     \rderiv = \rdyn(\rstate, \rctrl)
% \end{equation}
%
% The robot is operating an environment whose map is known. 
% In particular, let $\map$ represents the set of all obstacles in the environment.
% The goal of this paper is to determine a set of controls $\rctrl^{0:\finaltime}$ such that $\rstate^{\tvar} \not\in \hstate^{\tvar} \cup \map, \forall \tvar \in [0,\finaltime]$ and $\rstate^{\finaltime} = \rgoal$.
% Here, $\rgoal$ represents the goal state of the robot.
The robot's goal is to determine a set of controls $\rctrl^{0:\finaltime}$ such that it does not collide with the human or the (known) static obstacles, and reaches its goal $\rgoal$ by $\finaltime$.
In this work, we solve this planning problem in a receding horizon fashion.
Since the future states of the human are not known \textit{a priori}, we instead plan the robot trajectory to avoid $\frspos^{\tvar}_{\epsilon}(\cdot)$, the likely states of the human in the time interval $[\tvar, \tvar+\horizon]$, during planning at time $\tvar$.
% Even though conservative, this approach allows us to avoid the human at all times.
% \adnote{don't say it's conservative, because you can set your own epsilon}

\example{Our ground robot is modeled as a 3D Dubins' car with state given by its position and heading $\rstate = [s_x, s_y, \phi]$, and speed and angular speed as the control $\rctrl = [v_R, \omega]$. The respective dynamics are described by $\dot{x}_R = [v_R\cos\phi, v_R\sin\phi, \omega]$.
% %
% \begin{equation}
% \begin{aligned}
% \dot{s}_x = v_R\cos\phi,\quad \dot{s}_y = v_R\sin\phi,\quad \dot{\phi} = \omega\,, \\
% \underline{v} \le v_R \le \bar{v},\quad |\omega| \le \bar{\omega}.
% \end{aligned}
% \end{equation}
% %
At any given time $\tvar$, we use a third-order spline planner to compute the robot trajectory for a horizon of $\horizon$ (for more details, see~\cite{bansal2019combining}). 
%The spline trajectory is computed via an MPC scheme that avoids the static obstacles and the likely human positions, and navigates the robot towards the goal $\rgoal$.
%For the further details of the planner, we refer the interested readers to \cite{bansal2019combining}.
}
\section{Reachability-based Motion Prediction}
%In general, computing $\frspos^{\tvar}_{\epsilon}(\cdot)$ is a computationally challenging problem.
%Thereby, we propose a reachability-based framework to compute these sets.  
In this section, we first discuss how to cast the full probabilistic human motion prediction problem as a stochastic reachability problem.
Next, we discuss how we can obtain a deterministic approximation of this stochastic reachability problem, and solve it using our HJ reachability framework.
% and then discuss HJ reachability analysis, a framework to solve this reachability problem. \adnote{add something about the stochastic first, deterministic but over belief after}
% 
% \begin{itemize}
%     \item We formulate prediction over human state and $P(\param)$ as a reachability problem
%     \item We define a new joint state space and dynamics
%     \item This problem can be solved via some powerful numerical approaches
% \end{itemize}

\subsection{Casting prediction as a reachability problem}
\label{sec:stochastic_fromulation}
% \adnote{this is starting too abruptly, needs some setup}
% \adedit{We cast the problem of predicting a probability distribution over human states as some future time as one of maintaining a time-dependent distribution over reachable states and beliefs, given a stochastic human model as in (refer to eq 2).}
We cast the problem of predicting a probability distribution over the human states at some future time as one of maintaining a time-dependent distribution over reachable states and beliefs, given a stochastic human model as in \eqref{eq:control_policy}. Let the current time be $\tvar$ with the current (known) human state $\hstate^{\tvar}$ and a belief $\ptparam$. 
Different control actions $\hctrl^{\tvar}$ that the human might take next will induce a change in both the human's physical state and the robot's belief over $\param$. This in turn affects what human action distribution the robot will predict for the following timestep, and so on.
To simultaneously compute all possible future beliefs over $\param$ and corresponding likely human states, we consider the joint dynamics of $\ptparam$ and the human:
\begin{equation} \label{eq:joint_dynamics}
\dot{\jointstate}^{\tvar} = [\dot{x}_{H}^{\tvar}~~ \dot{\belief}^{\tvar}(\param)]=  \jointdyn({\jointstate}^{\tvar}, \hctrl^{\tvar}). 
                % = \begin{bmatrix}
                % \hdyn(\hstate^{\tvar}, \hctrl^{\tvar}) \\
                % \gamma \left(\postpt(\param | \hctrl^\tvar) - \ptparam\right) + \betaintrdyn(\ptparam)
                % \end{bmatrix}
\end{equation}
At any state $\jointstate$, the distribution over the (predicted) human actions is given by
\begin{equation} \label{eqn:jointdyn_controlpolicy}
    \hctrl \sim P(\hctrl \mid \jointstate) = \displaystyle\sum\nolimits_{\bar{\param}}P(\hctrl \mid \hstate; \bar{\param})\belief(\bar{\param}).
\end{equation}

To derive the dynamics of $\ptparam$ in \eqref{eq:joint_dynamics}, we note that the belief can change either due to the the new observations (via the Bayesian update in \eqref{eqn:bayesian_update}) or the change in human behavior (modeled via the parameter $\param$) over time. 
% there could be two sources of changes in the belief: (a) the new observations can potentially change this distribution via the Bayesian update (see \eqref{eqn:bayesian_update}), and (b) the human behavior (modeled via parameter $\param$) itself might change over time. 
This continuous evolution of $\ptparam$ can be described by the following equation:
%
% \begin{equation} \label{eqn:beta_dynamics}
%     \dot{\pt}(\param) = \gamma \left(\postpt(\param | \hctrl^\tvar) - \ptparam\right) + \betaintrdyn(\ptparam).
% \end{equation}
\begin{equation} \label{eqn:beta_dynamics}
    \dot{\belief}^{\tvar}(\param) = \gamma \Big(\postpt(\param \mid \hctrl^\tvar, \hstate^\tvar) - \ptparam\Big) + \betaintrdyn\Big(\ptparam\Big).
\end{equation}
Here, the function $\betaintrdyn$ represents the intrinsic changes in the human behavior, whereas the other component captures the Bayesian change in $\ptparam$ due to the observation $\hctrl^\tvar$. 
% \adnote{i'm not really familiar with this notation, not sure what $\gamma$ does}
Note that the time derivative in \eqref{eqn:beta_dynamics} is pointwise in the $\param$ space.

Typically, the Bayesian update is performed in discrete time when the new observations are received; however, in this work, we reason about continuous changes in $\ptparam$ and the corresponding continuous changes in the human state.
We omit a detailed derivation, but intuitively, to relate continuous-time Bayesian update to discrete-time version, $\gamma$ in \eqref{eqn:beta_dynamics} can be thought of as the observation frequency. 
Indeed, as $\gamma \uparrow \infty$, i.e., observations are received continuously, $\ptparam$ instantaneously changes to $\postpt(\param \mid \hctrl^\tvar, \hstate^\tvar)$. 
On the other hand, as $\gamma \downarrow 0$, i.e., no observation are received, the Bayesian update does not play a role in the dynamics of $\ptparam$.

Given the joint state at time $\tvar$, ${\jointstate}^{\tvar}$, and the control policy in \eqref{eqn:jointdyn_controlpolicy}, we are interested in computing the following set:
\begin{equation}
    \frs(\tdummy) = \{\jointstate^{\tdummy} \mid P(\jointstate^{\tdummy} \mid \jointstate^{\tvar}) > 0\}, \tdummy \in [\tvar, \tvar + \horizon].
\end{equation}
Intuitively, $\frs(\tdummy)$ represents all possible states of the joint system, i.e., all possible human states and beliefs over $\param$, that are reachable under the dynamics in \eqref{eq:joint_dynamics} for some sequence of human actions.
We refer to this set as \textit{Belief Augmented Forward Reachable Set (\frsname)} from here on.
Given a \frsname, we can obtain $\frspos^{\tvar}(\tdummy)$ by projecting $\frs(\tdummy)$ on the human state space. In particular,
\begin{equation}
    \frspos^{\tvar}(\tdummy) = \bigcup_{\jointstate^{\tdummy} \in \frs(\tdummy)} \proj(\jointstate^{\tdummy}),~~\tdummy \in [\tvar, \tvar + \horizon],
\end{equation}
where $\proj(\jointstate)$ denotes the human state component of $\jointstate$.
Consequently, the probability of any human state can be obtained as $P(\hstate^{\tdummy}) = \sum_{\jointstate^{\tdummy} \in \frs(\tdummy)} P(\jointstate^{\tdummy} \mid \jointstate^{\tvar}), \text{ if}~ \hstate^{\tdummy} = \proj(\jointstate^{\tdummy})$ (and $0$ otherwise) 
%
% \begin{equation}
%     P(\hstate^{\tdummy}) = 
%         \begin{cases}
%             \sum_{\jointstate^{\tdummy} \in \frs(\tdummy)} P(\jointstate^{\tdummy} \mid \jointstate^{\tvar}),& \text{if}~ \hstate^{\tdummy} = \proj(\jointstate^{\tdummy})\\
%             0,& \text{otherwise}, 
%         \end{cases}
% \end{equation}
%
which can be used to obtain $\frspos^{\tvar}_{\epsilon}(\tdummy)$ for any $\epsilon \ge 0$.
%
% \begin{remark}
% Even though theoretically powerful, the reachability formulation presented in this section, can be computationally prohibitive for real-world systems. 
% However, when $\param$ takes some small number of discrete values, modern computational tools can be used to solve this reachability problem, as we illustrate later in this paper.\adnote{this remark belongs in the deterministic framework!}
% \end{remark}

Since the control policy in \eqref{eqn:jointdyn_controlpolicy} is stochastic, the computation of $\frs(\tdummy)$ is a stochastic reachability problem.
However, when the prior $\ptparam$ is not correct, safeguarding against $\frs(\tdummy)$ is not sufficient.
Moreover, even when the prior is correct, computing stochastic reachable sets can be computationally demanding \cite{abate2007computational}.
To overcome these challenges, we recast the computation of $\frs(\tdummy)$ as a \textit{deterministic} reachability problem.
We next discuss HJ-reachability analysis for computing the \frsname{}
thanks to modern computational tools \cite{mitchell2004toolbox, chen2013flow}, and discuss how we can cast $\frs(\tdummy)$ as a deterministic reachability problem.

\begin{figure} [t!]
    \centering
    \includegraphics[width=\columnwidth]{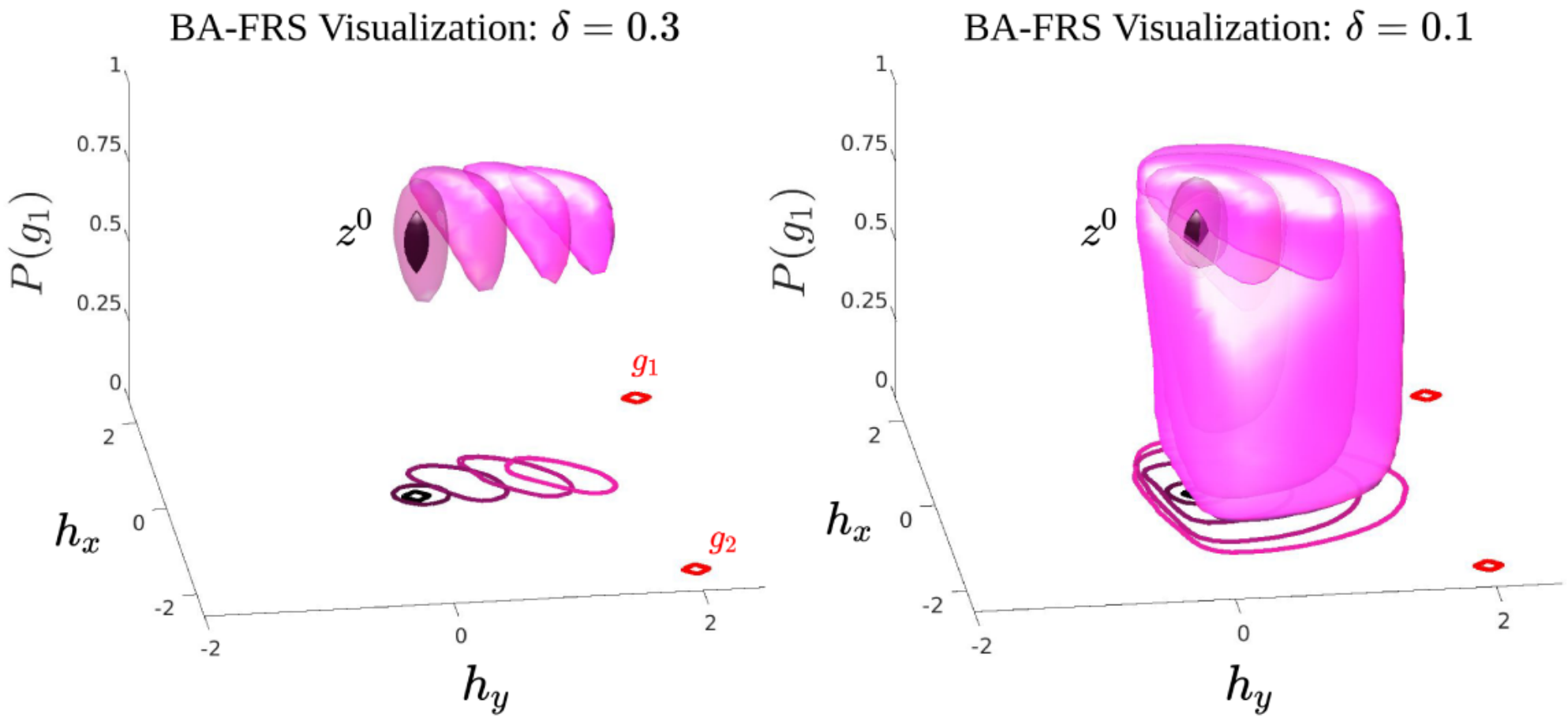}
    \caption{Visualization based on the the running example of the full BA-FRS and their projections into the human state space in the $h_x$-$h_y$ plane.}
    \label{fig:full_3D_sets}
    \vspace{-2em}
\end{figure}

\subsection{Background: Hamilton-Jacobi Reachability} \label{subsec:hj_reachability}
% \begin{itemize}
%     \item \abnote{put here the standard setup for reachability}
%     \item Recall that our goal is to determine where the human will be in the future. More formally, we want to compute the set of all states that our human dynamical system can reach from the initial set of states $\mathcal{L}$ at time $T$. 
%     \item This is a \textit{forward reachable set} computation and can be mathematically written as an intial-value PDE:
%     \begin{align}
%         \frac{\partial V}{\partial t} + \max_{\hctrl \in \mathcal{U}} \nabla_\hstate V(\hstate,t) \cdot \hdyn(\hstate,\hctrl) &= 0 \\
%         V(\hstate, 0) &= l(\hstate(0)) 
%     \end{align}
%     where $\mathcal{L} = \{\hstate: l(\hstate(0)) \leq 0\}$ where $\hstate(0)$ is the initial human state.
% \end{itemize}
% We use HJ reachability analysis to compute a forward reachable set (FRS) $\frs(\tdummy)$ given a set of starting states $\targetset$ \cite{lygeros2004reachability, mitchell2005time, bansal2017hamilton}.
% This analysis has been successfully applied in a variety of domains, such as aircraft auto-landing and safe large-scale multi-vehicle path planning \cite{bansal2017safe, bansal2017hamilton}.
HJ-reachability analysis \cite{lygeros2004reachability, mitchell2005time, bansal2017hamilton} can be used for computing a general Forward Reachable Set (FRS) $\frs(\tdummy)$ given a set of starting states $\targetset$.
Intuitively, $\frs(\tdummy)$ is the set of all possible states that the system can reach at time $\tdummy$ starting from the states in $\targetset$ under some permissible control sequence.
The computation of the FRS can be formulated as a dynamic programming problem which ultimately requires solving for the value function $\vfunc(\tdummy,\state)$ in the following initial value Hamilton Jacobi-Bellman PDE (HJB-PDE):
% %
% \begin{equation}
% \label{eq:HJB_FRS}
% \begin{aligned}
% &D_{\tdummy} \vfunc(\tdummy, \state) + \ham(\tdummy, \state, \nabla \vfunc(\tdummy, \state)) = 0, \\
% &\vfunc(0, \state) = \targetfunc(\state), \quad \tdummy \ge 0.
% \end{aligned}
% \end{equation}
% %
%
\begin{equation}
\label{eq:HJB_FRS}
D_{\tdummy} \vfunc(\tdummy, \state) + \ham(\tdummy, \state, \nabla \vfunc(\tdummy, \state)) = 0,~~ \vfunc(0, \state) = \targetfunc(\state),
\end{equation}
where $\tdummy \ge 0$. 
Here, $D_{\tdummy} \vfunc(\tdummy, \state)$ and $\nabla \vfunc(\tdummy, \state)$ denote the time and space derivatives of the value function respectively.
The function $\targetfunc(\state)$ is the implicit surface function representing the initial set of states $\targetset = \{\state: \targetfunc(\state) \le 0\}$.
The Hamiltonian, $\ham(\tdummy, \state, \nabla \vfunc(\tdummy, \state))$, encodes the role of system dynamics and control, and is given by
\begin{equation}
\label{eq:HJB_ham}
\ham(\tdummy, \state, \nabla \vfunc(\tdummy, \state)) = \max_{\hctrl \in \cset} \nabla \vfunc(\tdummy, \state) \cdot \jointdyn(\state, \hctrl).
\end{equation}

Once the value function $\vfunc(\tdummy, \state)$ is computed, the FRS is given by $\frs(\tdummy) = \{\state: \vfunc(\tdummy, \state) \le 0\}$.
% %
% \begin{equation*}
% \frs(\tdummy) = \{\state: \vfunc(\tdummy, \state) \le 0\}.
% \label{eq:frs_set}
% \end{equation*}
% % 
% HJ reachability also provides the optimal human action that maximally expands the FRS and is given by 
% %
% \begin{equation}
% \label{eq:opt_ctrl}
% \hctrl^{*}(\tdummy,\state) =  \arg \max_{\ctrl \in \cset} \min_{\dstb \in \dset} \nabla \vfunc(\tdummy, \state) \cdot \dyn(\state, \ctrl, \dstb).
% \end{equation}

%\subsection{HJ Reachability-based FRS Computation}
\subsection{An HJ Reachability-based framework for prediction \\ and analysis}
\label{subsec:hj_framework}

In this section, we build on the reachability formalism for prediction in Sec. \ref{sec:stochastic_fromulation} to obtain a framework which we will use to both generate robust and faster predictions, as well as to enable planners to answer important analysis questions about Bayesian predictors. 
Our framework is based on one key idea: rather than using a probability distribution over human actions as in \eqref{eqn:jointdyn_controlpolicy}, we will use a deterministic set of allowable human actions at every step. 
Very importantly, this set will be state-dependent, and therefore \emph{belief}-dependent:
\begin{equation} \label{eqn:allowable_control_general}
\hctrl \in \cset(z),\quad \cset(z) = \{\hctrl: h(\hctrl,z)\geq \delta \}
\end{equation}
where $h$ is a function allowed to depend on both the control and the state $\jointstate=(\hstate,\param)$, and threshold $\delta$.
% The control set depends on the threshold $\delta$, which is a design parameter.
Using a control set rather than a distribution allows us to convert the stochastic reachability problem in Sec. \ref{sec:stochastic_fromulation} to a deterministic reachability problem, which can be solved using the HJB-PDE formulation in Section \ref{subsec:hj_reachability}.
% 
% Different choices of $h$ in \eqref{eqn:allowable_control_general} will lead to different sets $\frs(\tdummy)$.
We now illustrate how different instantiations of $h$ in our framework enable both robust prediction and predictor analysis.

\noindent\textbf{Prediction.} We generate a predictor using our framework by instantiating the set of allowable human actions to be only those with sufficient probability under the belief: \begin{equation} \label{eqn:jointdyn_HJ_controlpolicy}
    \hctrl \in \cset(z),\quad \cset(z) = \{\hctrl: P(\hctrl \mid \jointstate) \ge \delta\}.
\end{equation}
Now, instead of associating future states with probabilities, we maintain a set of feasible states $z$ at every time step. Over time, this set still evolves via \eqref{eq:joint_dynamics}, but now all actions that have too low probability are excluded, and actions that have high probability are all treated as equally likely. However, because of the coupling between future belief and allowable actions, we may approximate $\frspos^{\tvar}_{\epsilon}(\tdummy)$ via a $\approxfrspos^{\tvar}_{\delta}(\tdummy)$, using a non-zero $\delta$.
% \propto \epsilon^{\frac{1}{\gamma\tdummy}}$. 
This has two advantages: (a) when the prior is correct, this allow us to compute an approximation of $\frspos^{\tvar}_{\epsilon}(\tdummy)$ using the computational tools developed for reachability analysis, and (b) when prior is incorrect, this allows the predictions to be robust to such inaccuracies, since computation of $\approxfrspos^{\tvar}_{\delta}(\tdummy)$ no longer relies on the exact action (or observation) probabilities.
We further discuss these aspects in Sec. \ref{sec:new_predictor}.
% In Sec. \ref{sec:new_predictor}, we demonstrate how $\approxfrspos^{\tvar}_{\delta}(\tdummy)$ relates to $\frspos^{\tvar}_{\epsilon}(\tdummy)$ empirically and discuss the relative pros and cons.

\noindent\textbf{Analysis.} Suppose we have a prior (or current belief) over $\param$; however, the prior might be wrong, i.e. $\arg\max_{\param^{'}} P(\param^{'})\neq \param^*$, with $\param^*$ being some hypothesized ground truth value for the human internal state. 
A reasonable question to ask in such a scenario would be ``how long it would take the robot to realize that the value of the internal state is $\param^*$, i.e. to place enough probability in its posterior on $\param^*$?'' 
A different instantiation of our framework can be used to answer such questions: we now want to compute the \frsname{} under allowed human actions that are modeling the hypothesized ground truth, and compute how long it takes to attain the desired property on the belief (we discuss this further in Sec. \ref{sec:predictor_analysis}). 
Thus, the allowed control set is:
\begin{equation} \label{eqn:jointdyn_HJ_controlpolicy_gt}
    \hctrl \in \cset(\jointstate),\quad \cset(\jointstate) = \{\hctrl: P(\hctrl \mid \hstate, \param^*) \ge \delta\}.
\end{equation}

Overall, by choosing $h$ appropriately, we can generate a range of predictors and analyses. 
The two examples above seemed particularly useful to us, and we detail them in the following sections. 
\section{A New HJ Reachability-based Predictor} \label{sec:new_predictor}
Our reachability-based framework enables us generate a new predictor by computing an approximation of \frsname{}.
% using the allowable human controls set for a future joint state $\jointstate$ to be those with high enough probability given $\jointstate$. 
In this section, we analyze the similarities and differences between this predictor and the one obtained by solving the full stochastic reachability problem.

\begin{figure} [t!]
    \centering
    \includegraphics[width=0.9\columnwidth]{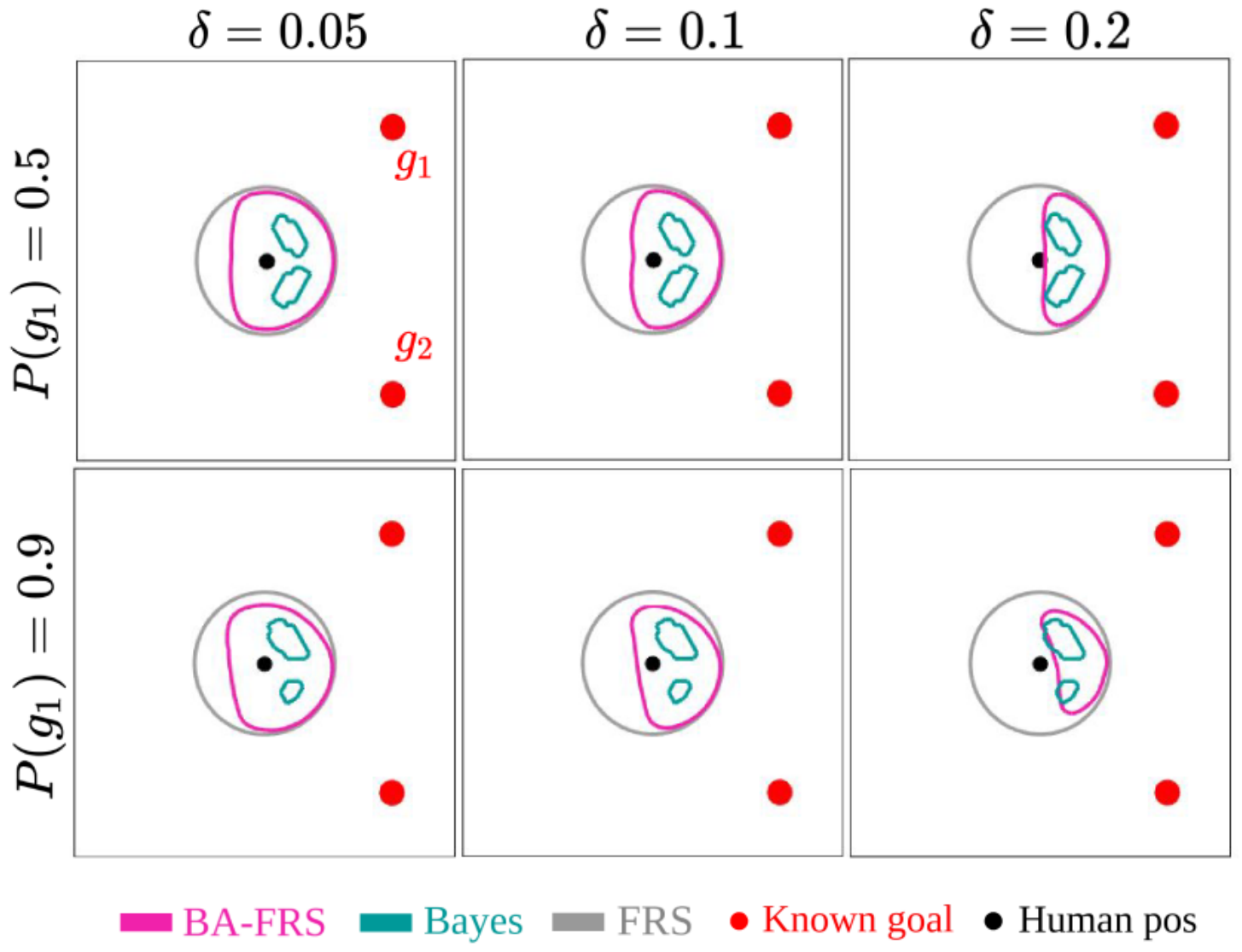}
    \caption{As $\delta$ increases, we obtain a tighter approximation of $\frspos^{\tvar}$; this is because a sequence of ``medium'' likely actions over time result in states that are overall unlikely under a Bayesian prediction. Discarding such actions under our framework leads to a better approximation of $\frspos^{\tvar}$. However, choosing $\delta$ too aggressively might lead to an overly optimistic set.}
    \label{fig:frs_illustration}
    \vspace{-2em}
\end{figure}

\noindent\textbf{Prediction as an approximation of $\frspos$.}
Since the stochastic reachability problem needs to explicitly maintain the state probabilities, it might be challenging to compute $\frspos$ compared to $\approxfrspos$.
% Since the human motion often needs to be predicted at run time, $\approxfrspos$ thus allows for a more scalable prediction. 
However, this advantage in computation complexity is achieved at the expense of losing the information about the human state distribution, which can be an important component for several robotic applications.
However, when the full state distribution is not required, as is the case in this paper, $\approxfrspos$ provides a very good approximation of $\frspos$. 
In fact, it can be shown that $\frspos^{\tvar}_0(\tdummy) = \approxfrspos^{\tvar}_0(\tdummy)$.
% \adnote{explain here how it relates, use running example}
\example{
% When $\epsilon$ and $\delta$ are non-zero, the theoretical relationship between the two sets depends on the human policy and the initial belief; however, we now empirically demonstrate that $\approxfrspos^{\tvar}(\tdummy)$ still provides a good approximation of $\frspos^{\tvar}(\tdummy)$.
For simplicity, consider the case when the intrinsic behavior of the human does not change over time, i.e., $\betaintrdyn(\ptparam) = 0$. 
% We also let $\gamma = 1$.
Since in this case $\param$ takes only two possible values, the joint state space is three dimensional. 
In particular, $\jointstate = [h_x, h_y, p_1]$, where $p_1 := \belief(\param = g_1)$.
$\belief(\param = g_2)$ is given by $\left(1-\belief(\param = g_1)\right)$ so we do not need to explicitly maintain it as a state.
$P(\hctrl \mid \jointstate) = p_1 \mathcal{N}(\mu_1, \sigma_1^2) + (1 - p_1)\mathcal{N}(\mu_2, \sigma_2^2)$
%$P(\hctrl \mid \jointstate)$ is given by
%
%\begin{equation*}
%    P(\hctrl \mid \jointstate) = \frac{p_1}{\sqrt{2\pi\sigma^2_1}} e^{-\frac{(\hctrl - \mu_1)^2}{2\sigma^2_1}} + \frac{(1-p_1)}{\sqrt{2\pi\sigma^2_2}} e^{-\frac{(\hctrl - \mu_2)^2}{2\sigma^2_2}},
%\end{equation*}
%
which can be used to compute the set of allowable controls $\cset$ for different $\delta$s as per \eqref{eqn:jointdyn_HJ_controlpolicy}. 
% Recall that $\mu_1$ and $\mu_2$ are dependent on the human state, and hence the set of allowable controls is state dependent. 
% However, our reachability framework can easily handle this scenario.
We use the Level Set Toolbox \cite{mitchell2004toolbox} to compute the \frsname{}, starting from $\hstate = (0, 0)$.
The corresponding likely human states, $\approxfrspos^{\tvar}_{\delta}(\horizon)$, for different initial priors and $\delta$s are shown in Fig. \ref{fig:frs_illustration} in magenta.
For comparison purposes, we compute $\frspos^{\tvar}_{\epsilon}(\horizon)$ (teal), as well as the ``naive'' FRS obtained using all possible human actions (gray).
$\epsilon$ for $\frspos$ is picked to capture 95\% area of the set. \\
As evident from Fig. \ref{fig:frs_illustration}, $\approxfrspos^{\tvar}$ is an over-approximation of $\frspos^{\tvar}$, but at the same time it is not overly conservative unlike the naive FRS. 
This is primarily because even though the proposed framework doesn't maintain the full state distribution, it still discards the unlikely controls during the \frsname{} computation.
It is also interesting to note that \frsname{} is not too sensitive to the initial prior for low $\delta$s. 
This property of \frsname{} allows the predictions to be robust to incorrect priors as we explain later in this section. 
\\
We also show the full 3D \frsname{}, as well as the projected $\approxfrspos^{\tvar}$ sets over time for an initial prior, $p_1 = 0.8$, in Fig. \ref{fig:full_3D_sets}.
When $\delta$ is high, both the belief as well as the human states are biased towards the goal $g_1$ over time. 
When $\delta$ is high, only the actions that steer the human towards $g_1$ will be initially contained in the control set for the \frsname{} computation.
Moreover, propagating the current belief under these actions further reinforces the belief that the human is moving towards $g_1$.
As a result, the beliefs in the \frsname{} shift towards a higher $p_1$ over time.
On the other hand, when $\delta=0.1$, the human actions under $g_2$ are also contained in the control set, which leads to the belief and the human state shift in both directions (towards $g_1$ and $g_2$). 
% However, there is still more propagation towards $g_1$, as expected given the initial prior.
}

\noindent\textbf{Prediction as robust to incorrect priors and misspecified models.} 
% \adnote{stuff about misspecification}
% In this section, we illustrate some of the additional advantages of the proposed reachability-based prediction scheme.
% \subsection{Robustness to Incorrect Priors and Misspecified Models}
The set $\frspos^{\tvar}_{\epsilon}(\horizon)$ depends heavily on the prior $\ptparam$.
% Moreover, the corresponding $\epsilon$ that one needs to pick also depends heavily on this prior to make sure that $\frspos^{\tvar}(\cdot)$ is not too conservative or optimistic. 
When the initial prior for the human motion prediction is not accurate enough, using $\frspos^{\tvar}_{\epsilon}(\horizon)$ for planning might lead to unsafe behavior as it can be too optimistic.
This issue is particularly exacerbated when the true (unknown) parameter of the human, $\param^{*}$, is not within the support of $\param$ considered in the model, i.e., when the model is misspecified.
For example, when the exact goal of the human may not be any of the goals specified in the model.
In such scenarios, a full Bayesian inference may fail to assign sufficient probabilities to the likely states of the human, which can lead to unsafe situations.
On the other hand, using the \textit{full FRS}, i.e., the set of all states the human can reach under \textit{any} possible control, will ensure safety but can impede robot plan efficiency. 

In such situations, the proposed framework presents a good middle-ground alternative to the two approaches -- it does not rely heavily on the exact probability of an action while computing FRS since it leverages action probabilities only to distinguish between likely and unlikely actions.
Yet, it still uses a threshold to discard highly unlikely actions under the current belief, ensuring the obtained FRS is not too conservative.
This allows our framework to perform well in situations where the initial prior is not fully accurate but accurate enough to distinguish likely actions from unlikely actions.
In particular, suppose the prior at time $\tvar$ is such that $P(\hctrl \mid \hstate^\tvar; \param^{*}) \ge \delta \implies P(\hctrl \mid \jointstate^\tvar) \ge \delta ~\forall \hctrl$, 
% %
% \begin{equation} \label{eqn:robustness_condition}
% P(\hctrl \mid \hstate^\tvar; \param^{*}) \ge \delta \implies P(\hctrl \mid \jointstate^\tvar) \ge \delta ~\forall \hctrl,
% \end{equation}
% %
where $\param^{*}$ is the true (unknown) human parameter, and $\jointstate^\tvar = (\hstate^\tvar, \ptparam)$.
Intuitively, the above condition states that the prior at time $\tvar$ is accurate enough to distinguish the set of likely actions from the unlikely actions for the \textit{true} human behavior; however, we do not have the knowledge of the true probability distribution of the actions. 
In such a case, it can be shown that any human state that is reachable under a control sequence consisting of at least $\delta$-likely controls will be contained within $\approxfrspos^{\tvar}_{\delta}$. 

\begin{figure} [t!]
    \centering
    \includegraphics[width=0.8\columnwidth]{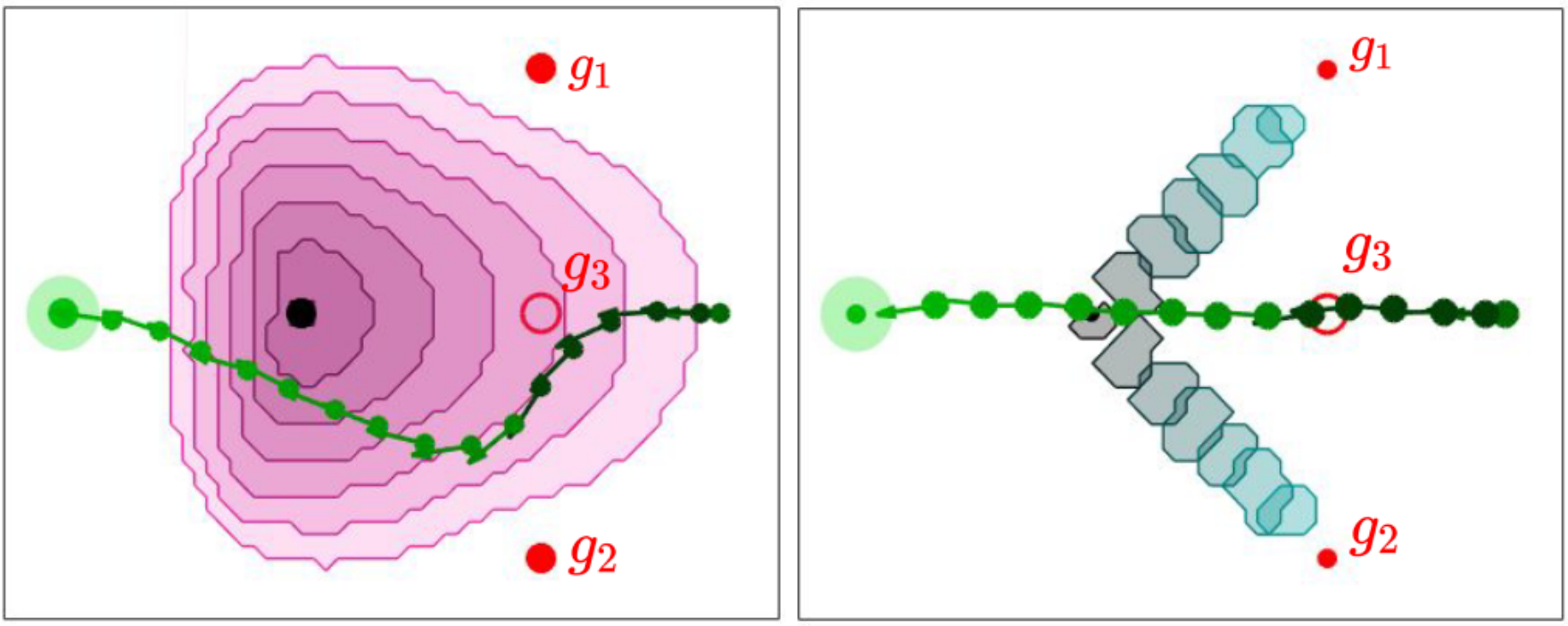}
    \caption{Visualized is a snapshot of the human predictions $\frspos^{\tvar}_{\epsilon}(\horizon)$ starting from a prior of $(0.5, 0.5)$, and the corresponding trajectory of the robot when using the BA-FRS (left) or the Bayesian predictor (right).}
    \label{fig:misspecified_goal}
    \vspace{-2em}
\end{figure}

\example{Consider the scenario where the actual human goal is $g_3$, midway between $g_1$ and $g_2$ (see Fig. \ref{fig:misspecified_goal}). 
Thus, the current model does not capture the true goal of the human.
Even though the human walks straight towards $g_3$, a full Bayesian framework fails to assign sufficient probabilities to the likely human states because of its over reliance on the model.
Ultimately, this leads to a collision between the human and the robot.
In contrast, since our framework uses the model to only distinguish likely actions from unlikely actions, it recognizes that moving straight ahead is a likely human action.
This is also evident from Fig. \ref{fig:frs_illustration}, where the states straight ahead of the human are contained within the \frsname{} even for a relatively high $\delta$ of 0.2.
As a result, using the deterministic \frsname{} for planning leads to a safe navigation around the human. \\
These results are confirmed in our hardware experiments performed on a TurtleBot 2 testbed. As shown in Fig.~\ref{fig:front_fig}, we demonstrate these scenarios for navigating around a human participant. We measured human positions at ~200Hz using a VICON motion capture system and used on-board TurtleBot odometry sensors for the robot state measurement. As discussed, our framework allows us to be robust to misspecified goals while not being overly conservative. 
}

\section{Prediction Analysis} \label{sec:predictor_analysis}
% \adnote{modified into section, tweaked title}
% \adnote{i suggested an edit above where you would put the equation under that section (C), so remember to remove it from here and point back to that}
%One of the main objectives of Bayesian inference is to determine the true behavior of the human from a series of observations starting from a prior.
%Thus, one interesting question to ask is how much time (or equivalently the number of observations) will be required to determine the true human behavior under the current prior? Since our reachability based framework explicitly maintain all possible beliefs that are reachable under the current belief, it is well equipped to answer such questions.
% 
An interesting question that our framework can answer is how long does the predictor have to observe the human in order to determine the true human behavior for some prior.
For simplicity, consider the scenario where $\param$ can take two possible values $b_1$ or $b_2$; however, the true human parameter is unknown.
We also have an initial prior over $\param$, given by $\belief^0(\param)$.
% Without loss of generality, suppose the true human parameter $\param^* = b_1$. 
Then we may pose the following questions: (1) What is the minimum and maximum possible time it will take to determine that $\param^* = b_1$ with sufficiently high probability (denoted as $\tmin^1$ and $\tmax^1$)? (2) What are the corresponding sequences of observations?
%two specific questions that could of interest:
%\begin{enumerate}
%    \item What is the minimum and maximum possible time it will take to determine that $\param^* = b_1$ with sufficiently high probability (denoted as $\tmin^1$ and $\tmax^1$)?
%    \item What are the corresponding sequences of observations? 
%\end{enumerate}
Thus, we want to know what is the most and the least informative set of observations that the human can provide under $\param^* = b_1$.
Similar questions can also be posed for $b_2$.
The overall minimum and the maximum time to determine the true human behavior are then given by $\tmin = \min\{\tmin^1, \tmin^2\}$ and $\tmax = \min\{\tmax^1, \tmax^2\}$.
Once $\tmin$ and $\tmax$ are available, the robot trajectory can be planned to safeguard against both possibilities ($\param^* = b_1$ and $\param^* = b_2$) for $\tvar \le \tmax$. 
After a duration of $\tmax$, we will be able to determine the true human behavior so it is sufficient to safeguard against the likely states of the human under the belief $\belief^{\tmax}(\param)$ there on.

As discussed in Sec. \ref{subsec:hj_framework}, an instantiation of the proposed reachability-based framework can be used to determine $\tmin$ and $\tmax$. 
In particular, given the initial human state $\hstate^0$ and the initial prior $\belief^0(\param)$, we can compute the \frsname{} with the control policy in \eqref{eqn:jointdyn_HJ_controlpolicy_gt},
% %
% \begin{equation}
%     \hctrl \in \cset,\quad \cset = \{\hctrl: P(\hctrl \mid \hstate; \beta_1) \ge \delta\},
% \end{equation}
% %
and the Hamiltonian $\ham(\tdummy, \state, \nabla \vfunc(\tdummy, \state)) = \max_{\hctrl \in \cset} \nabla \vfunc(\tdummy, \state) \cdot \jointdyn(\state, \hctrl).$
% 
% \begin{equation}
% \ham(\tdummy, \state, \nabla \vfunc(\tdummy, \state)) = \max_{\hctrl \in \cset} \nabla \vfunc(\tdummy, \state) \cdot \jointdyn(\state, \hctrl).
% \end{equation}
Then, $\tmin^1$ can be obtained as the minimum time such that $P^*_1(\param)$ is contained in $\frs(\cdot)$ for some human state $\hstate$.
Here, $P^*_1(\param)$ is a distribution that assigns a sufficiently high probability to $\param = b_1$.
Intuitively, this computation gives the minimum time it will take to reach the belief that $\param^* = b_1$ if the true human parameter is indeed $b_1$ and the human is giving us the most informative samples to discern its behavior.  
We can similarly compute $\tmin^2$ by computing a similar \frsname{} under the likely controls from $b_2$.

Similarly, $\tmax^1$ can be computed using a similar procedure, but instead of maximizing over the control in the Hamiltonian, we minimize over the control instead. 
This computation corresponds to finding the control sequence that is least informative at inferring $\param^* = b_1$ and can be obtained by: $
\hctrl^{*}(\tdummy,\jointstate) =  \arg \min_{\hctrl \in \cset} \nabla \vfunc(\tdummy, \jointstate) \cdot \jointdyn(\jointstate, \hctrl).
$
% 
% This computation corresponds to finding the set of joint states that the system will reach even when the human applies the control to minimize the growth of the \frsname{}.
% As a result, at $\tmax^1$, all possible trajectories that start from $\hstate^0$ and $\belief^0(\param)$ will reach the belief $P^*_1(\param)$ regardless of the control sequence applied by the human. 
% Consequently, within a duration of $\tmax^1$, we will be able to infer that the true human behavior is modeled by $b_1$, if that is indeed the case.
% The control sequence that is least informative at inferring $\param^* = b_1$ (and hence corresponds to the last state to reach the belief $P^*_1(\param)$) can be obtained as:
% %
% \begin{equation}
% \label{eq:opt_ctrl}
% \hctrl^{*}(\tdummy,\jointstate) =  \arg \min_{\hctrl \in \cset} \nabla \vfunc(\tdummy, \jointstate) \cdot \jointdyn(\jointstate, \hctrl).
% \end{equation}
%
Intuitively, $\hctrl^{*}$ is the control observation that least differentiate between $b_1$ and $b_2$. 
Similarly, $\hctrl^{*}$ corresponding to the computation of $\tmin$ is the control observation that differentiates most between $b_1$ and $b_2$. 
This is closely related to prior work on legibility \cite{dragan2013legibility} and deception \cite{dragan2014analysis}: given a fixed horizon, our framework computes a sequence of controls that is maximally informative or maximally ambiguous cumulatively across all the time steps, which in general is nontrivial to compute.
% Interestingly, the most informative sequence of actions for one hypothesis will not always coincide to the sequence of highest probability actions under that hypothesis -- this is the distinction between acting legibly and acting predictably -- which makes this informative sequence and the $\tmin$ nontrivial to compute.

\example{
Consider the planar pedestrian dynamics as before, but with the following human policy:
\begin{equation}
       \hctrl^{\tvar} \mid \hstate^{\tvar} \sim 
    \begin{cases}
        \mathcal{N}(0, \sigma^2),  & \text{if } \param = 0 \\
        \text{Uniform}(-\pi, \pi),  & \text{if } \param = 1,
    \end{cases} 
\end{equation}
where $\sigma = 0.1$. 
The human walks straight with a small variance when $\param = 0$ and move in a random direction when $\param = 1$, approximating an irrational human.
We compute the minimum and maximum time to realize $\lambda^* = 0$ starting from a high initial prior on irrational behavior, $(0.1, 0.9)$.
We assume that we can confidently conclude that $\lambda^* = 0$ when all human trajectories reach a belief of at least $0.9$ for $\lambda = 0$.
The obtained $\tmin$ and $\tmax$ are 3.2s and 11.2s respectively.
We also compute the control sequences that correspond to these times. 
The optimal control sequence for $\tmin$ is given by $\hctrl = 0$, since that is the most likely action under the rational behavior compared to irrational behavior.
On the other hand, for $\tmax$, the optimal control sequences consist of an angle of 15 degrees, which is the least likely action that is above the $\delta$-threshold (0.3 for this example) for $\lambda = 0$.
}

\section{Conclusion}% and Future Work}
When robots operate in complex environments around humans, they often employ probabilistic predictive models to reason about human behavior.
Even though powerful, such predictors can make poor predictions when the prior is incorrect or the observation model is misspecified. This in turn will likely cause unsafe behavior.
% While a full probabilistic prediction maybe tractable in some cases, it can be extremely difficult for agents whose intent and preferences are evolving over time.
In this work, we formulate human motion  prediction as a Hamilton-Jacobi reachability problem. 
We demonstrate that the proposed framework provides more robust predictions when the prior is incorrect or human behavior model is misspecified, can perform these predictions in continuous time and state using the tools developed for reachability analysis, and can be used for the predictor analysis. 
%Extending our methodology to multi-robot, multi-human settings as well as to other online learning-based settings are exciting future directions. %as well as to other application spaces such as manipulation.
% 
% and then present a Hamilton-Jacobi reachability-based framework that not only can compute an approximation of likely human states at a significant lower computational complexity, but can also perform predictor analysis.
% We demonstrate that the proposed framework also provides more robust predictions when the human behavior model is inaccurate, and demonstrate our approach in simulation and hardware.
% Extending our methodology to multi-robot, multi-human settings as well as to other online learning-based settings are exciting future directions. %as well as to other application spaces such as manipulation.

% \bibliographystyle{plainnat}
% \bibliography{references}
\newpage
% \balance
\printbibliography

@article{lygeros2004reachability,
  title={On reachability and minimum cost optimal control},
  author={Lygeros, John},
  journal={Automatica},
  volume={40},
  number={6},
  pages={917--927},
  year={2004},
  publisher={Elsevier}
}

@inproceedings{bansal2017hamilton,
  title={{Hamilton-Jacobi} Reachability: A Brief Overview and Recent Advances},
  author={Bansal, S. and Chen, M. and Herbert, S. and Tomlin, C. J.},
  booktitle={CDC},
  year={2017},
}

@article{mitchell2005time,
  title={A time-dependent Hamilton-Jacobi formulation of reachable sets for continuous dynamic games},
  author={Mitchell, I. and Bayen, A. and Tomlin, C. J.},
  journal={IEEE Trans. on automatic control},
  volume={50},
  number={7},
  pages={947--957},
  year={2005},
}

@article{mitchell2004toolbox,
  title={A toolbox of level set methods},
  author={Mitchell, I.},
  journal={http://www. cs. ubc. ca/mitchell/ToolboxLS/toolboxLS. pdf, Tech. Rep. TR-2004-09},
  year={2004}
}

@article{bansal2019combining,
  title={Combining Optimal Control and Learning for Visual Navigation in Novel Environments},
  author={Bansal, S. and Tolani, V. and Gupta, S. and Malik, J. and Tomlin, C.},
  journal={arXiv preprint},
  year={2019}
}

@article{fisac2018general,
  title={A general safety framework for learning-based control in uncertain robotic systems},
  author={Fisac, J. and Akametalu, A. and Zeilinger, M. and Kaynama, S. and Gillula, J. and Tomlin, C. J.},
  journal={IEEE Trans. on Automatic Control},
  year={2018},
}

@article{rudenko2019human,
  title={Human Motion Trajectory Prediction: A Survey},
  author={Rudenko, Andrey and Palmieri, Luigi and Herman, Michael and Kitani, Kris M and Gavrila, Dariu M and Arras, Kai O},
  journal={arXiv preprint arXiv:1905.06113},
  year={2019}
}

@inproceedings{lasota2017multiple,
  title={A multiple-predictor approach to human motion prediction},
  author={Lasota, Przemyslaw A and Shah, Julie A},
  booktitle={2017 IEEE International Conference on Robotics and Automation (ICRA)},
  pages={2300--2307},
  year={2017},
  organization={IEEE}
}

@inproceedings{ziebart2009planning,
  title={Planning-based prediction for pedestrians},
  author={Ziebart, Brian D. and Ratliff, Nathan and Gallagher, Garratt and Mertz, Christoph and Peterson, Kevin and Bagnell, J. Andrew and Hebert, Martial and Dey, Anind K and Srinivasa, Siddhartha},
  booktitle={International Conference on Intelligent Robots and Systems (IROS)},
  pages={3931--3936},
  year={2009},
}

@inproceedings{chen2013flow,
  title={Flow*: An analyzer for non-linear hybrid systems},
  author={Chen, Xin and {\'A}brah{\'a}m, Erika and Sankaranarayanan, Sriram},
  booktitle={International Conference on Computer Aided Verification},
  pages={258--263},
  year={2013},
  organization={Springer}
}

@article{driggs2018robust,
  title={Robust, Informative Human-in-the-Loop Predictions via Empirical Reachable Sets},
  author={Driggs-Campbell, Katherine and Dong, Roy and Bajcsy, Ruzena},
  journal={IEEE Transactions on Intelligent Vehicles},
  year={2018},
  publisher={IEEE}
}

@article{kochenderfer2010airspace,
  title={Airspace encounter models for estimating collision risk},
  author={Kochenderfer, Mykel J and M. Edwards, Matthew W and Espindle, Leo P and Kuchar, James K and Griffith, J Daniel},
  journal={Journal of Guidance, Control, and Dynamics},
  volume={33},
  number={2},
  pages={487--499},
  year={2010}
}

@inproceedings{ng2000algorithms,
  title={Algorithms for inverse reinforcement learning.},
  author={Ng, Andrew Y and Russell, Stuart J and others},
  booktitle={International Conference on Machine Learning (ICML)},
  pages={663--670},
  year={2000}
}

@inproceedings{bai2015intention,
  title={Intention-aware online POMDP planning for autonomous driving in a crowd},
  author={Bai, Haoyu and Cai, Shaojun and Ye, Nan and Hsu, David and Lee, Wee Sun},
  booktitle={International Conference on Robotics and Automation (ICRA)},
  pages={454--460},
  year={2015},
}

@inproceedings{baker2007goal,
  title={Goal inference as inverse planning},
  author={Baker, Chris L and Tenenbaum, Joshua B and Saxe, Rebecca R},
  booktitle={Annual Meeting of the Cognitive Science Society},
  volume={29},
  year={2007}
}

@incollection{bandyopadhyay2013intention,
  title={Intention-aware motion planning},
  author={Bandyopadhyay, Tirthankar and Won, Kok Sung and Frazzoli, Emilio and Hsu, David and Lee, Wee Sun and Rus, Daniela},
  booktitle={Algorithmic Foundations of Robotics X},
  pages={475--491},
  year={2013},
  publisher={Springer}
}

@inproceedings{amor2014interaction,
  title={Interaction primitives for human-robot cooperation tasks},
  author={Amor, Heni Ben and Neumann, Gerhard and Kamthe, Sanket and Kroemer, Oliver and Peters, Jan},
  booktitle={International Conference on Robotics and Automation (ICRA)},
  pages={2831--2837},
  year={2014},
}

@inproceedings{ding2011human,
  title={Human arm motion modeling and long-term prediction for safe and efficient human-robot-interaction},
  author={Ding, Hao and Rei{\ss}ig, Gunther and Wijaya, Kurniawan and Bortot, Dino and Bengler, Klaus and Stursberg, Olaf},
  booktitle={International Conference on Robotics and Automation},
  pages={5875--5880},
  year={2011},
}

@inproceedings{koppula2013anticipating,
  title={Anticipating human activities for reactive robotic response.},
  author={Koppula, Hema Swetha and Saxena, Ashutosh},
  booktitle={International Conference on Intelligent Robots and Systems},
  pages={2071},
  year={2013}
}

@article{lasota2015analyzing,
  title={Analyzing the effects of human-aware motion planning on close-proximity human--robot collaboration},
  author={Lasota, Przemyslaw A and Shah, Julie A},
  journal={Human factors},
  volume={57},
  number={1},
  pages={21--33},
  year={2015},
  publisher={Sage Publications Sage CA: Los Angeles, CA}
}

@inproceedings{hawkins2013probabilistic,
  title={Probabilistic human action prediction and wait-sensitive planning for responsive human-robot collaboration},
  author={Hawkins, Kelsey P and Vo, Nam and Bansal, Shray and Bobick, Aaron F},
  booktitle={International Conference on Humanoid Robots (Humanoids)},
  pages={499--506},
  year={2013},
}

@article{Schmerling2017,
archivePrefix = {arXiv},
arxivId = {1710.09483},
  title={Multimodal Probabilistic Model-Based Planning for Human-Robot Interaction},
  author={Schmerling, Edward and Leung, Karen and Vollprecht, Wolf and Pavone, Marco},
  journal={arXiv preprint arXiv:1710.09483},
  year={2017}
}

@inproceedings{finn2016guided,
  title={Guided cost learning: Deep inverse optimal control via policy optimization},
  author={Finn, Chelsea and Levine, Sergey and Abbeel, Pieter},
  booktitle={International Conference on Machine Learning},
  pages={49--58},
  year={2016}
}

@inproceedings{kitani2012activity,
  title={Activity forecasting},
  author={Kitani, Kris M and Ziebart, Brian D and Bagnell, James Andrew and Hebert, Martial},
  booktitle={European Conference on Computer Vision},
  pages={201--214},
  year={2012},
}

@InProceedings{Ma_2017_CVPR,
author = {Ma, Wei-Chiu and Huang, De-An and Lee, Namhoon and Kitani, Kris M.},
title = {Forecasting Interactive Dynamics of Pedestrians With Fictitious Play},
booktitle = {Conference on Computer Vision and Pattern Recognition (CVPR)},
month = {July},
year = {2017}
}

@inproceedings{alahi2016social,
  title={Social lstm: Human trajectory prediction in crowded spaces},
  author={Alahi, Alexandre and Goel, Kratarth and Ramanathan, Vignesh and Robicquet, Alexandre and Fei-Fei, Li and Savarese, Silvio},
  booktitle={Conference on Computer Vision and Pattern Recognition (CVPR)},
  pages={961--971},
  year={2016}
}

@inproceedings{sadigh2016information,
 author = {Sadigh, Dorsa and Sastry, S. Shankar and Seshia, Sanjit A. and Dragan, Anca},
 title = {Information Gathering Actions over Human Internal State},
 booktitle = {Proceedings of the {IEEE},
 /{RSJ},
 International Conference on Intelligent Robots and Systems (IROS)},
 pages = {66--73},
 year = {2016},
 month = {October},
 publisher = {IEEE},
 doi = {10.1109/IROS.2016.7759036}
}

@article{abate2008probabilistic,
  title={Probabilistic reachability and safety for controlled discrete time stochastic hybrid systems},
  author={Abate, Alessandro and Prandini, Maria and Lygeros, John and Sastry, Shankar},
  journal={Automatica},
  volume={44},
  number={11},
  pages={2724--2734},
  year={2008},
  publisher={Elsevier}
}

@inproceedings{abate2007computational,
  title={Computational approaches to reachability analysis of stochastic hybrid systems},
  author={Abate, Alessandro and Amin, Saurabh and Prandini, Maria and Lygeros, John and Sastry, Shankar},
  booktitle={International Workshop on Hybrid Systems: Computation and Control},
  pages={4--17},
  year={2007},
  organization={Springer}
}

@inproceedings{dragan2013legibility,
  title={Legibility and predictability of robot motion},
  author={Dragan, Anca D and Lee, Kenton CT and Srinivasa, Siddhartha S},
  booktitle={Proceedings of the 8th ACM/IEEE international conference on Human-robot interaction},
  pages={301--308},
  year={2013},
  organization={IEEE Press}
}

@inproceedings{dragan2014analysis,
  title={An Analysis of Deceptive Robot Motion.},
  author={Dragan, Anca D and Holladay, Rachel M and Srinivasa, Siddhartha S},
  booktitle={Robotics: science and systems},
  pages={10},
  year={2014},
  organization={Citeseer}
}

\end{document}